\documentclass[preprint,12pt,authoryear]{elsarticle}

\usepackage{graphicx}%
\usepackage{multirow}%
\usepackage{amsmath,amssymb,amsfonts}%
\usepackage{amsthm}%
\usepackage{xcolor}%
\usepackage{textcomp}%
\usepackage{booktabs}%
\usepackage{bm}%
\usepackage{subcaption}%

\usepackage[pagewise, mathlines]{lineno}

\usepackage[colorlinks]{hyperref}
\hypersetup{
  citecolor = {blue}
}

\usepackage{geometry}
\theoremstyle{thmstyleone}%
\theoremstyle{thmstyletwo}%
\theoremstyle{thmstylethree}%

\begin{document}

\begin{frontmatter}

\title{Regime-Conditional Stabilisation of LLM-Augmented Cooperative Multi-Agent Reinforcement Learning}

\author[ensia]{Faid Keddouri\corref{cor1}}
\ead{faid.keddouri@ensia.edu.dz}

\author[ensia]{Sohaib Houhou}

\author[ensia]{Aissa Boulmerka}

\author[uge]{Nadir Farhi}

\cortext[cor1]{Corresponding author.}

\address[ensia]{National School of Artificial Intelligence (ENSIA), Sidi Abdellah Campus, Algiers, Algeria}

\address[uge]{Cosys-Grettia, Univ Gustave Eiffel, F-77454 Marne-la-Vallee, France.}

\begin{abstract}
Large Language Models (LLMs) offer a natural interface for translating
human objectives into reward signals for cooperative multi-agent reinforcement
learning (MARL), yet the training-time dynamics of this integration remain poorly
understood. We show that dynamically updating LLM-generated reward weights during
off-policy MARL violates the stationarity assumption of Potential-Based Reward
Shaping (PBRS) and contaminates the experience replay buffer, whose stored
transitions carry reward labels computed under stale shaping weights. We
characterise the result as a \emph{regime-dependent} failure whose severity depends
on how competent the unshaped baseline already is. To control it we propose two
stabilisation strategies: a Phase-Based Freeze Schedule that enforces strict
stationarity within training phases, and Exponential Moving Average (EMA) smoothing
that bounds per-episode weight drift. We evaluate across three cooperative
environments and five random seeds with QMIX, complemented by an exploratory VDN
extension, yielding a three-regime taxonomy. In the \emph{augmentative} regime
(Simple Spread), where the baseline is functional (74.4\%), EMA significantly
improves success to 86.7\% ($+12.3$\,pp, $p<0.01$) while naive dynamic updates
collapse it to 15.2\%. In the \emph{essential} regime (Level-Based Foraging), where
the baseline is broken (0.1\%), any shaping unlocks the task (95.9\% under EMA). In
the \emph{supplementary} regime (SMAC 3m), where the baseline is near-saturated
(98.8\%), stabilised shaping preserves performance (99.9\%) while unstabilised
shaping adds variance without gain. These findings establish reward-signal
stationarity as a necessary design constraint and indicate that regime placement
is a practical predictor of whether dynamic LLM shaping helps or harms.
\end{abstract}


\end{frontmatter}

\section{Introduction}\label{sec:intro}

Cooperative multi-agent reinforcement learning (MARL) has produced strong team
behaviour in complex coordination tasks through value-decomposition methods such
as QMIX \citep{rashid2018qmix} and policy-gradient methods such as MAPPO
\citep{yu2022mappo}. A persistent practical obstacle, however, is adapting a
learned team policy to \emph{shifting human objectives} without exhaustive
retraining. Reward shaping is the natural lever: a human expresses a preference,
and that preference is injected as an auxiliary learning signal. Yet manual reward
design is notoriously difficult --- hand-crafted rewards are error-prone,
task-specific, and inaccessible to non-expert operators --- which has motivated a
rapidly growing line of work that places a Large Language Model (LLM) at this
interface. LLMs have been used as proxy reward functions scored directly from
textual task descriptions \citep{kwon2023reward}, as generators of reward code for
robotic skill synthesis \citep{yu2023language}, and as iterative reward designers
--- Eureka \citep{ma2024eureka}, Text2Reward \citep{xie2024text2reward}, and the
multi-agent LAMARL \citep{zhu2025lamarl} translate natural-language instructions
into executable reward signals. These systems operate predominantly in single-agent
or on-policy settings and do not examine what happens when shaping weights are
\emph{updated mid-training} inside an off-policy learner with experience replay.

The dynamic setting is not a corner case: it is the natural deployment mode of a
language interface to reward. The very appeal of expressing objectives in natural
language is that they can be \emph{revised} --- an operator monitoring a team of
agents will want to shift emphasis from speed to safety, or from individual
progress to global coverage, as circumstances change, without restarting a training
run that may span days. Existing frameworks sidestep this by treating reward
generation as an offline step that concludes before learning begins (or between
entire training runs, in Eureka's evolutionary loop). The moment weight updates
move \emph{inside} the training run, however, they interact with the machinery of
off-policy learning, and that interaction has not been characterised.

This omission is consequential. Value-decomposition methods are off-policy by
construction: their sample efficiency rests on an experience replay buffer that
reuses transitions collected thousands of episodes earlier. The buffer stores
transitions whose reward labels were computed under the shaping
weights in force at collection time. When an LLM revises those weights during
training, stored rewards become inconsistent with the current objective: a
transition collected early is later sampled with a reward that no longer reflects
the prevailing reward function. Potential-Based Reward Shaping (PBRS)
\citep{ng1999policy} guarantees that shaping of the form
$F(s,s') = \gamma\Phi(s') - \Phi(s)$ preserves the optimal policy, but only while
the potential $\Phi$ is \emph{stationary}. An LLM that re-issues weights
periodically makes $\Phi$ time-varying, breaking this guarantee precisely in the
algorithmic setting --- off-policy replay --- where the resulting inconsistency is
most damaging.

We study this failure systematically and find that its severity is not uniform but
\emph{regime-dependent}: it is governed by how competent the unshaped baseline
policy already is. This paper makes three contributions.

\begin{itemize}
  \item \textbf{(C1) A three-regime taxonomy} of cooperative environments ---
  \emph{essential}, \emph{augmentative}, and \emph{supplementary} --- that predicts
  whether dynamic LLM reward shaping unlocks a task, risks catastrophic collapse,
  or is redundant but harmless once stabilised. With one exemplar environment per
  regime, we advance the taxonomy as an organising hypothesis supported by three
  case studies rather than as an established law.
  \item \textbf{(C2) Two stabilisation strategies} that restore stationarity to the
  shaped potential: a \emph{Phase-Based Freeze Schedule} enforcing strict
  stationarity within training phases, and \emph{Exponential Moving Average (EMA)}
  smoothing bounding per-episode weight drift, with the smoothing factor selected by
  ablation ($\alpha=0.2$).
  \item \textbf{(C3) A multi-condition empirical validation} across three
  cooperative environments (Simple Spread, Level-Based Foraging, SMAC 3m) and five
  random seeds with QMIX, complemented by an exploratory VDN extension that probes
  whether the taxonomy is specific to QMIX's monotonic mixer.
\end{itemize}

The remainder of the paper is organised as follows. Section~\ref{sec:related}
reviews related work; Section~\ref{sec:background} introduces the formal background;
Section~\ref{sec:method} presents the language-to-reward pipeline, the
non-stationarity analysis, and the two stabilisation strategies;
Section~\ref{sec:experiments} reports the multi-environment results and the regime
taxonomy; Section~\ref{sec:interp} covers interpretability and human-in-the-loop
evaluation; Section~\ref{sec:discussion} discusses implications and future directions;
and Section~\ref{sec:conclusion} concludes.

\section{Related Work}\label{sec:related}

\paragraph{Cooperative MARL and value decomposition.} Centralised Training with
Decentralised Execution (CTDE) within the Decentralised Partially Observable Markov
Decision Process (Dec-POMDP) framework \citep{oliehoek2016concise} underpins modern
cooperative MARL. On the policy-gradient side, MADDPG \citep{lowe2017maddpg}
introduced centralised critics with decentralised actors, and MAPPO
\citep{yu2022mappo} showed that carefully tuned on-policy learning is competitive
across standard benchmarks. On the value-based side, VDN \citep{sunehag2018vdn}
decomposes the joint action-value additively, QMIX \citep{rashid2018qmix} factors it
through a monotonic mixing network, and later work relaxes the monotonicity
restriction --- QTRAN \citep{son2019qtran} via transformed factorisation objectives
and QPLEX \citep{wang2021qplex} via a duplex dueling architecture. Progress in this
family has been driven substantially by standardised benchmarks such as the
StarCraft Multi-Agent Challenge \citep{samvelyan2019smac}, from which our SMAC 3m
task is drawn. \emph{Gap:} these methods assume a fixed team reward and do
not study the interaction between a dynamically shifting reward function and the
off-policy transitions stored under earlier objectives.

\paragraph{Potential-Based Reward Shaping.} Ng et al. \citep{ng1999policy} proved
that shaping rewards of the form $F = \gamma\Phi(s') - \Phi(s)$ leave the optimal
policy invariant, \emph{provided the potential $\Phi$ is stationary}, i.e.\ held
fixed throughout learning. Static potential-based shaping was later shown to be
equivalent to initialising the $Q$-function with $\Phi$
\citep{wiewiora2003equivalence}, and the framework was extended so that arbitrary
reward functions can be expressed as potential-based advice
\citep{harutyunyan2015advice}.
Devlin and Kudenko \citep{devlin2012dynamic} extended the invariance
guarantee to time-varying potentials in \emph{on-policy} cooperative settings, where
fresh interactions remain consistent with the current potential. \emph{Gap:} this
guarantee does not transfer to off-policy algorithms with experience replay, where
stored transitions carry reward labels from earlier potentials and the telescoping
sum that underlies policy invariance no longer cancels.

\paragraph{LLMs as reward designers.} An LLM was first used as a proxy reward
function that scores agent behaviour directly against a textual task description
\citep{kwon2023reward}; subsequent work translates language into reward code for
real-time robotic skill synthesis \citep{yu2023language}. Eureka \citep{ma2024eureka}
couples LLM reward generation with evolutionary search over full reward programs,
and Text2Reward \citep{xie2024text2reward} generates dense reward code refined by
human feedback, both for single-agent tasks; LAMARL \citep{zhu2025lamarl} extends
LLM-aided reward design to cooperative MARL. All of these systems revise rewards
\emph{between} training runs or before training begins, and they measure success by
the semantic quality of the generated reward. \emph{Gap:} none of these frameworks
characterise the replay-buffer contamination that arises when LLM-produced reward
weights shift mid-training in an off-policy learner, nor do they constrain LLM
outputs to a stationarity-preserving potential.

\paragraph{Non-stationarity in multi-agent learning.} Standard MARL
non-stationarity arises from concurrently evolving policies: each agent's
environment shifts as its teammates learn \citep{hernandezleal2017survey}.
Experience replay can be stabilised against this policy-version drift by
conditioning on a training-iteration fingerprint \citep{foerster2017stabilising} or
by importance-weighting stale transitions. A separate line, Reinforcement Learning
from Human Feedback (RLHF), learns reward models from human preference comparisons
\citep{christiano2017deep} and now underpins the alignment of large language models
themselves \citep{ouyang2022training}; there the reward model also evolves, but
training is typically on-policy and the reward is inferred implicitly rather than
issued as explicit weights. \emph{Gap:} these approaches address
\emph{policy-version} non-stationarity or implicit reward inference; the failure
studied here is \emph{reward-label} non-stationarity injected by explicit external
LLM weight updates, orthogonal to policy-version awareness.

\section{Background}\label{sec:background}

Table~\ref{tab:notation} collects the notation used throughout the paper; each
symbol is also defined in the text at its first occurrence.

\begin{table}[!tbp]
  \caption{Table of notations.}
  \label{tab:notation}
  \centering
  \begin{tabular}{@{}ll@{}}
    \toprule
    Symbol & Meaning \\
    \midrule
    $G$ & Dec-POMDP tuple $\langle \mathcal{I},\mathcal{S},\mathcal{A},P,R,\Omega,O,\gamma\rangle$ \\
    $\mathcal{I}$, $n$ & set of agents and their number \\
    $\mathcal{S}$, $\mathcal{A}$ & state space and joint action space $A_1\times\cdots\times A_n$ \\
    $P$, $R$ & transition function and shared team reward \\
    $\Omega$, $O$ & observation space and observation function \\
    $\gamma$ & discount factor \\
    $o_t^{(i)}$, $u_t^{(i)}$ & observation and action of agent $i$ at step $t$ \\
    $\tau_i$, $\pi_i$ & action-observation history and policy of agent $i$ \\
    $Q_i$, $Q_{\mathrm{tot}}$ & per-agent utility and joint action-value \\
    $\Phi$, $F$ & shaping potential and PBRS reward $\gamma\Phi(s')-\Phi(s)$ \\
    $r_{\mathrm{base}}$, $r_{\mathrm{shaped}}$ & unshaped team reward $R(s,\bm{u})$ and shaped reward $r_{\mathrm{base}}+F$ \\
    $I$ & natural-language operator instruction \\
    $\bm{\phi}(s)$, $m$ & feature-function vector $(\phi_1,\dots,\phi_m)^\top$ and its dimension ($m=5$) \\
    $\bm{w}$, $\bm{w}_{\mathrm{LLM}}$ & shaping weight vector in force and raw LLM output \\
    $\alpha$ & EMA smoothing factor \\
    $K$, $P_k$, $\bm{w}^{(k)}$ & number of phases, $k$-th phase, and its frozen weights \\
    $T$ & training horizon in episodes \\
    $\eta$, $\varepsilon$ & learning rate and exploration rate \\
    $\sigma$ & cross-seed standard deviation \\
    \bottomrule
  \end{tabular}
\end{table}

\paragraph{Dec-POMDP.} A cooperative multi-agent task is modelled as a Decentralised
Partially Observable Markov Decision Process \citep{oliehoek2016concise}
$G=\langle \mathcal{I}, \mathcal{S}, \mathcal{A}, P, R, \Omega, O, \gamma \rangle$,
where $\mathcal{I}$ is the set of $n$ agents, $\mathcal{S}$ the state space,
$\mathcal{A}=A_1\times\cdots\times A_n$ the joint action space, $P$ the transition
function, $R:\mathcal{S}\times\mathcal{A}\to\mathbb{R}$ the shared team reward,
$\Omega$ the observation space with observation function $O$, and
$\gamma\in[0,1)$ the discount factor. Agents share a common reward but each observes
only part of the environment; agent $i$ conditions its policy
$\pi_i(u_i\mid\tau_i)$ on its local action-observation history $\tau_i$. The team
maximises the expected discounted return
$J(\pi)=\mathbb{E}\big[\sum_{t=0}^{\infty}\gamma^t r_t\big]$.

\paragraph{Value decomposition.} QMIX \citep{rashid2018qmix} learns per-agent
utilities $Q_i(\tau_i,u_i)$ and combines them through a monotonic mixing network
into a joint action-value $Q_{\mathrm{tot}}$, enforcing the
Individual-Global-Max condition via the constraint
$\partial Q_{\mathrm{tot}}/\partial Q_i \ge 0$ so that decentralised greedy action
selection is jointly optimal. VDN \citep{sunehag2018vdn} is the special case
$Q_{\mathrm{tot}}=\sum_i Q_i$. Both are off-policy and trained from an experience
replay buffer.

\paragraph{Potential-Based Reward Shaping.} PBRS incorporates domain knowledge
without altering the optimal policy. Ng et al. \citep{ng1999policy} proved that a
shaping reward
\begin{equation}\label{eq:pbrs}
F(s,s') = \gamma\,\Phi(s') - \Phi(s)
\end{equation}
added to the environment reward leaves the optimal policy unchanged for any
real-valued potential $\Phi:\mathcal{S}\to\mathbb{R}$, because its discounted
contribution telescopes to a policy-independent constant determined by the initial
state; they further showed that this potential-based form is \emph{necessary} for
policy invariance under arbitrary dynamics. The guarantee, however, requires $\Phi$
to be \emph{fixed} throughout learning. In MARL, PBRS likewise preserves Nash
equilibria \citep{devlin2012dynamic}, but again only under a stationary $\Phi$. As
Section~\ref{sec:method} shows, LLM-produced dynamic weights make $\Phi$
time-varying and violate this assumption.

\section{Methodology}\label{sec:method}

We present the methodology in four parts: the language-to-reward pipeline that turns
an instruction into a shaped reward (Section~\ref{sec:l2r}), the optimised QMIX
backbone (Section~\ref{sec:backbone}), a mechanistic account of the non-stationarity
that dynamic weight updates induce (Section~\ref{sec:nonstat}), and the two
stabilisation strategies that control it (Section~\ref{sec:stab}).

\subsection{Language-to-Reward Pipeline}\label{sec:l2r}

The pipeline is built on a fixed vector of $m=5$ bounded, interpretable feature
functions
$\bm{\phi}(s)=(\phi_1(s),\dots,\phi_m(s))^\top$, each mapping the state to a scalar
that captures one behavioural aspect of the task. Instantiated on Simple Spread,
the five features are:
\begin{itemize}
  \item $\phi_1$ --- \emph{local proximity}: how close each agent is to its nearest
  landmark, rewarding individual progress towards coverage;
  \item $\phi_2$ --- \emph{global coverage}: how well the team as a whole covers all
  landmarks, capturing the joint objective;
  \item $\phi_3$ --- \emph{collision avoidance}: a penalty term that decreases the
  potential when agents crowd within collision distance of one another;
  \item $\phi_4$ --- \emph{formation}: the spatial dispersion of the team,
  encouraging agents to spread out rather than cluster;
  \item $\phi_5$ --- \emph{speed}: the agents' movement magnitude, allowing an
  operator to trade urgency against safety.
\end{itemize}
The other environments instantiate the same five slots with domain analogues
(e.g.\ proximity to food and joint-load readiness in LBF, target proximity and
focus-fire concentration in SMAC 3m); the features are hand-designed once per
environment and never change during training.

\begin{figure}[!tbp]
  \centering
  \includegraphics[width=0.92\textwidth]{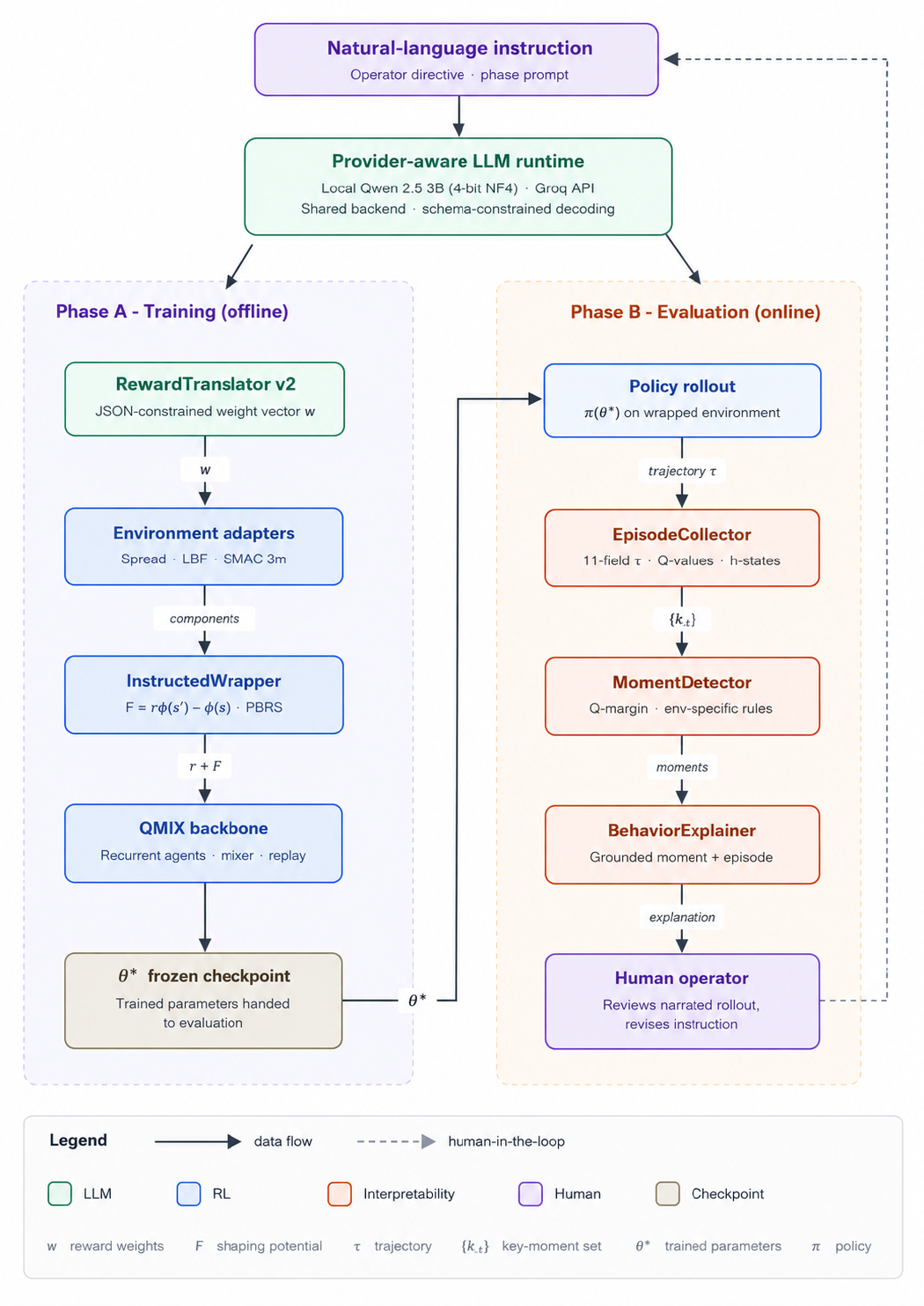}
  \caption{The provider-aware two-tier Language-to-Reward pipeline.
  \emph{Phase A (training, offline):} a natural-language instruction is translated by
  the LLM runtime into a JSON weight vector that parameterises the PBRS potential
  $\Phi$, which shapes rewards for the QMIX backbone and yields a frozen checkpoint
  --- no gradient flows back to the LLM. \emph{Phase B (evaluation, online):} the
  frozen policy is rolled out and an interpretability stack (EpisodeCollector
  $\to$ MomentDetector $\to$ BehaviorExplainer) produces natural-language audit
  narratives for a human operator.}
  \label{fig:pipeline}
\end{figure}

An LLM acts as a ``Reward Architect'': given a high-level natural-language
instruction $I$ (e.g. \emph{``avoid collisions and focus on global coverage''}), it
outputs a structured weight vector with one component per feature of
$\bm{\phi}$,
\begin{equation}\label{eq:weights}
\bm{w}_{\mathrm{LLM}} = \mathrm{LLM}(I) = \{ w_{\mathrm{local}}, w_{\mathrm{global}},
w_{\mathrm{collision}}, w_{\mathrm{formation}}, w_{\mathrm{speed}} \},
\end{equation}
whose $m=5$ components form the weight vector $\bm{w}=(w_1,\dots,w_m)^\top$, so the
named components of Equation~\eqref{eq:weights} correspond one-to-one to
$\phi_1,\dots,\phi_5$ above. Each weight is constrained to $w_j\in[0,5]$: a bounded
range caps the magnitude of the potential and hence of the shaping signal ---
together with the bounded features it keeps
$|F|/|r_{\mathrm{base}}|\approx 0.05$ (Section~\ref{sec:nonstat}), so shaping can
never dominate the base reward --- while still spanning a factor-of-five contrast
between de-emphasised and emphasised features. The coincidence between the upper
bound 5 and the number of features $m=5$ is incidental. As the reward architect we
use Qwen 2.5 3B (Instruct) \citep{qwen2024report} in 4-bit NF4 quantisation. The
choice is driven by the nature of the task rather than by benchmark performance:
translating a short instruction into a constrained JSON schema requires
instruction-following, not open-ended reasoning, so a small open-weights model
suffices; quantised, it occupies $\approx 1.91$\,GB of VRAM alongside training and
answers in under 200\,ms per query (Table~\ref{tab:llm}), which makes even
per-episode querying feasible. The LLM's role is strictly translational, decoupled
from the RL gradient loop; robustness to the choice of reward architect is flagged
as future work (Section~\ref{sec:discussion}). The weight vector parameterises a
potential
\begin{equation}\label{eq:potential}
\Phi(s;\bm{w}) = \sum_{j=1}^{m} w_j\,\phi_j(s) = \bm{w}^\top \bm{\phi}(s).
\end{equation}
The complete information flow is
\begin{equation}\label{eq:flow}
I \;\xrightarrow{\;\mathrm{LLM}\;}\; \bm{w}_{\mathrm{LLM}}
\;\longrightarrow\; \Phi(\,\cdot\,;\bm{w})
\;\longrightarrow\; F(s,s')
\;\longrightarrow\; r_{\mathrm{shaped}} = r_{\mathrm{base}} + F(s,s'),
\end{equation}
where $r_{\mathrm{base}}=R(s,\bm{u})$ denotes the unshaped team reward returned by
the environment and $F$ is the PBRS reward of Equation~\eqref{eq:pbrs}.
Figure~\ref{fig:pipeline} illustrates the pipeline: the MARL training loop produces
shaped rewards and updates policies through TD-error gradient steps, while the LLM
module receives an instruction and returns a JSON weight vector. Crucially, no
gradient signal flows back to the LLM --- it remains frozen throughout training.

\subsection{Optimised Recurrent QMIX Backbone}\label{sec:backbone}

We build on QMIX with parameter-shared Gated Recurrent Unit (GRU)
\citep{cho2014gru} agent networks. To resolve the perceptual
aliasing that arises in symmetric environments --- where agents receiving identical
observations cannot differentiate their roles --- we augment each agent's input with
a one-hot agent identity and its previous action:
\begin{equation}\label{eq:input}
\mathrm{Input}_t^{(i)} = \big[\, o_t^{(i)},\ \mathrm{OneHot}(u_{t-1}^{(i)}),\
\mathrm{OneHot}(i) \,\big].
\end{equation}
This enrichment raised the unshaped Simple Spread success rate from 55\% to
approximately 71\% in early single-seed development (1{,}000 evaluation episodes),
in line with QMIX benchmarks for this task \citep{papoudakis2021benchmarking},
though evaluation protocols differ (we use 200-episode greedy evaluation at each
checkpoint), so the comparison is indicative rather than exact. The canonical
five-seed baseline reported later (Figure~\ref{fig:mainresults}) is
$74.4\pm5.4$\%. All subsequent experiments use this backbone.

\subsection{Non-Stationarity: A Mechanistic Account}\label{sec:nonstat}

Under standard PBRS the shaping reward of Equation~\eqref{eq:pbrs} preserves optimal
policies when $\Phi$ is stationary. When the LLM re-issues weights during training,
the potential becomes time-dependent, $\Phi_t(s)=\bm{w}_t^\top\bm{\phi}(s)$ with
$\bm{w}_t\neq\bm{w}_{t+1}$, and the per-step shaping reward is
$F_t(s_t,s_{t+1})=\gamma\,\Phi_t(s_{t+1})-\Phi_t(s_t)$. The telescoping sum that
underlies policy invariance then fails: in the discounted shaping return
$\sum_t\gamma^t F_t$, the term $\gamma^{t+1}\Phi_t(s_{t+1})$ contributed at step $t$
no longer cancels the term $-\gamma^{t+1}\Phi_{t+1}(s_{t+1})$ contributed at step
$t+1$, leaving an uncancelled residual
$\gamma^{t+1}[\Phi_t(s_{t+1})-\Phi_{t+1}(s_{t+1})]$ at every interior state, so the
shaped Markov decision process is no longer reward-equivalent to the original.

This violation compounds through four interacting mechanisms.
\begin{itemize}
  \item \textbf{(i) PBRS guarantee violation.} With a stationary potential, the
  cumulative shaping reward telescopes to a policy-independent boundary term, so no
  policy is preferred over another on account of the shaping. Once $\Phi$ becomes
  time-varying, the uncancelled residuals derived above accumulate along
  trajectories, introducing a systematic bias that can steer the learned policy away
  from the optimum of the original task.
  \item \textbf{(ii) Replay-buffer contamination.} Off-policy learners store shaped
  rewards, not the quantities needed to recompute them. A transition stored at
  episode 1{,}000 carries a shaped reward computed with $\bm{w}_{1000}$; when it is
  sampled later under $\bm{w}_{5000}$, the stored label no longer reflects the
  prevailing objective. With a 5{,}000-episode buffer (Table~\ref{tab:qmix}), every
  minibatch mixes reward semantics from up to 5{,}000 episodes of weight history, so
  gradients average over mutually inconsistent objectives.
  \item \textbf{(iii) Moving target.} Even setting the buffer aside, per-episode
  weight updates shift the reward function faster than the value function can adapt
  to it: agents chase a moving objective rather than converging on any single one,
  and improvements credited under one weight vector may be penalised under the
  next.
  \item \textbf{(iv) Value-function instability.} TD targets embed the shaping term
  $F$. When $F$ changes between the moment a target is computed and the moment the
  corresponding gradient is applied --- an interval stretched further by the hard
  target-network updates every 200 episodes --- the regression targets themselves
  become noisy, inflating value-estimation variance on top of the bias from (i) and
  (ii).
\end{itemize}
A shaping-magnitude analysis confirms $|F|/|r_{\mathrm{base}}|\approx 0.05$, ruling out
reward domination: the failure arises from non-stationarity, not from the shaping
signal overwhelming the base reward.

We are explicit about the scope of this account: it identifies the mechanisms at
work and orders their interaction, but it does not quantify them. Deriving a formal
bound on TD-target bias as a function of weight drift and replay-buffer age --- the
off-policy counterpart of the on-policy time-varying-potential guarantee of Devlin
and Kudenko \citep{devlin2012dynamic} --- is an open theoretical problem; the empirical study of
Section~\ref{sec:experiments} tests the account's qualitative predictions instead.

\subsection{Stabilisation Strategies}\label{sec:stab}

We propose two complementary strategies that restore stationarity during off-policy
training, differing in the stationarity--adaptability trade-off.

\paragraph{Phase-Based Freeze Schedule.} We partition the $T$-episode horizon into
$K$ contiguous phases $P_1, P_2, \dots, P_K$, where phase $P_k$ covers episodes
$t\in\big[(k{-}1)\,T/K,\ k\,T/K\big)$. At the boundary into phase $P_k$ the LLM
produces $\bm{w}^{(k)}=\mathrm{LLM}(I_k)$ from the instruction $I_k$ active at that
point; within the phase, weights are frozen,
$\bm{w}_t=\bm{w}^{(k)}\ \forall t\in P_k$. This
guarantees a stationary $\Phi$ and exact PBRS invariance \emph{within} each phase, at
the cost of abrupt transitions between phases.

\paragraph{Exponential Moving Average (EMA) smoothing.} EMA relaxes strict
stationarity in favour of quasi-stationarity. Instead of applying
$\bm{w}_{\mathrm{LLM}}$ directly, we blend each new target with the previous weights:
\begin{equation}\label{eq:ema}
\bm{w}_t = \alpha\cdot\bm{w}_{\mathrm{LLM}} + (1-\alpha)\cdot\bm{w}_{t-1},
\end{equation}
with $\alpha\in(0,1)$. Subtracting $\bm{w}_{t-1}$ from Equation~\eqref{eq:ema} gives
$\bm{w}_t-\bm{w}_{t-1}=\alpha(\bm{w}_{\mathrm{LLM}}-\bm{w}_{t-1})$, so the per-episode
weight drift satisfies
$\|\bm{w}_t-\bm{w}_{t-1}\| = \alpha\|\bm{w}_{\mathrm{LLM}}-\bm{w}_{t-1}\|$, where
$\|\cdot\|$ is the Euclidean norm. Since
$\Phi_t(s)-\Phi_{t-1}(s)=(\bm{w}_t-\bm{w}_{t-1})^\top\bm{\phi}(s)$, the
Cauchy--Schwarz inequality bounds the induced potential drift, measured in the
sup-norm $\|\Phi_t-\Phi_{t-1}\|_\infty = \max_s|\Phi_t(s)-\Phi_{t-1}(s)|$:
\begin{equation}\label{eq:bound}
\|\Phi_t-\Phi_{t-1}\|_\infty \ \le\ \alpha\,\|\bm{w}_{\mathrm{LLM}}-\bm{w}_{t-1}\|
\cdot \max_s \|\bm{\phi}(s)\|,
\end{equation}
which approaches zero as $\alpha\to 0$. Equation~\eqref{eq:bound} controls the drift
of the potential itself, not the downstream value-function bias; it should be read
as a design principle --- bounded drift --- rather than a performance guarantee. We
fix $\alpha=0.2$ for all multi-seed experiments; Section~\ref{sec:alpha} justifies
this choice.

\section{Experiments and Results}\label{sec:experiments}

\subsection{Experimental Setup}\label{sec:setup}

We evaluate on three cooperative environments spanning a wide range of baseline
competence: Simple Spread (MPE), Level-Based Foraging (LBF), and SMAC 3m. Each method
is trained for 100{,}000 episodes across five random seeds $\{42,101,123,456,999\}$,
and we report the mean $\pm$ standard deviation. Evaluation uses a deterministic
greedy policy ($\varepsilon=0$) over 200 episodes at each checkpoint. The reported
metric is binary success rate for Simple Spread (all $N$ landmarks covered) and LBF
($\ge 1$ food collected through valid joint loads), and win rate for SMAC 3m. Because
MARL training exhibits seed variance, numbers should be compared \emph{within} an
environment, not across environments. We compare four conditions: Baseline (no shaping), Dynamic
LLM (per-episode weight updates), and the two stabilised variants (Freeze-Schedule
and EMA $\alpha=0.2$). All four share the backbone and training budget of
Table~\ref{tab:qmix}; the LLM configuration is given in Table~\ref{tab:llm} and the
three environments in Table~\ref{tab:envs}.

\paragraph{The Dynamic LLM condition.} In the Dynamic LLM condition the reward
architect is re-queried at every episode boundary, so the weight vector in force can
change from one episode to the next. Two sources of variation are therefore present:
scheduled revisions of the operator instruction over the course of training, and
per-query sampling stochasticity of the LLM (temperature 0.1) under the instruction
active at that point.
Per-episode re-querying is deliberately the most aggressive schedule: no existing
framework updates this frequently (Eureka revises rewards between training runs,
LAMARL at coarser granularity), and we do not present it as a reproduction of any
deployed system. It is instead the \emph{limiting case} of the setting that
motivates this work --- real-time human-in-the-loop steering, in which an operator
may revise the objective at any moment --- and thus an upper bound on the update
frequencies such a system could exhibit. The mechanistic account of
Section~\ref{sec:nonstat} predicts that intermediate update frequencies (e.g., every
few thousand episodes, without smoothing) interpolate between the Dynamic and
Freeze-Schedule endpoints, and that collapse severity scales with replay-buffer age;
testing these two causal predictions is left to future work
(Section~\ref{sec:discussion}).

\begin{table}[!tbp]
  \caption{Canonical QMIX configuration (shared across all conditions).}
  \label{tab:qmix}
  \centering
  \begin{tabular}{@{}ll@{}}
    \toprule
    Hyperparameter & Value \\
    \midrule
    Agent controller       & FC $\to$ GRU $\to$ FC \\
    GRU hidden dim         & 64 \\
    Mixer embedding dim    & 32 \\
    Optimiser              & Adam \\
    Learning rate          & $5\times10^{-4}$ \\
    Discount factor $\gamma$ & 0.99 \\
    Gradient clip          & 10.0 \\
    Batch size             & 32 episodes \\
    Replay buffer          & 5{,}000 episodes \\
    Target update          & Hard, every 200 episodes \\
    Exploration $\varepsilon$ & $1.0 \to 0.05$ over 20k ep. \\
    Training budget        & 100{,}000 episodes \\
    Eval protocol          & Greedy, 200 ep. / checkpoint \\
    \bottomrule
  \end{tabular}
\end{table}

\begin{table}[!tbp]
  \caption{Canonical LLM (Reward Architect) configuration.}
  \label{tab:llm}
  \centering
  \begin{tabular}{@{}ll@{}}
    \toprule
    Parameter & Value \\
    \midrule
    Model                   & Qwen 2.5 3B (Instruct) \\
    Quantisation            & 4-bit NF4, fp16 compute \\
    VRAM footprint          & $\approx 1.91$\,GB \\
    Translation latency     & $\approx 200$\,ms / query \\
    Translation temperature & 0.1 \\
    Weight range            & $w_j \in [0,5]$ \\
    Output                  & JSON-only weight vector \\
    \bottomrule
  \end{tabular}
\end{table}

\begin{table}[!tbp]
  \caption{The three evaluated environments span the full range of baseline
  competence.}
  \label{tab:envs}
  \centering
  \begin{tabular}{@{}llll@{}}
    \toprule
    Property & Simple Spread & LBF 8$\times$8-3p-2f & SMAC 3m \\
    \midrule
    Backend       & PettingZoo MPE & \texttt{lbforaging} v3 & SMAClite \\
    Agents        & 3 & 3 & 3 (Marines) \\
    Observation   & 18-dim local & 15-dim local & 30-dim local \\
    Actions       & Discrete(5) & Discrete(6) & Discrete(9) \\
    Horizon       & 25 steps & 50 steps & $\sim$30 steps \\
    Reward        & Dense, per-step & Sparse on food load & Sparse: damage $+$ win \\
    Cooperation   & Implicit (collisions) & Explicit (level-sum) & Implicit (focus-fire) \\
    Regime        & Augmentative & Essential & Supplementary \\
    SR/WR$_{\mathrm{base}}$ & 74.4\% & $\approx 0$\% & 98.8\% \\
    \bottomrule
  \end{tabular}
\end{table}

\subsection{Selecting the Smoothing Factor}\label{sec:alpha}

\begin{figure}[!tbp]
  \centering
  \includegraphics[width=0.95\textwidth]{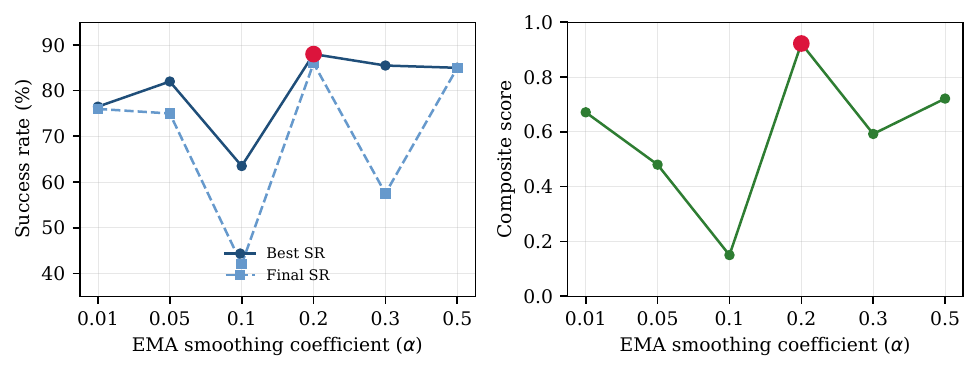}
  \caption{Diagnostic EMA $\alpha$-ablation landscape on Simple Spread.
  \emph{Left:} best and final success rates. \emph{Right:} the composite ranking
  statistic used only for within-sweep model selection. The landscape is
  non-monotonic; $\alpha=0.2$ (red marker) is jointly best on peak success, final
  success, return, and composite score.}
  \label{fig:alpha}
\end{figure}

The value $\alpha=0.2$ is not arbitrary. Before the multi-seed campaign, a diagnostic
single-seed sweep on Simple Spread (same 100k-episode budget) evaluated
$\alpha\in\{0.01,0.05,0.1,0.2,0.3,0.5\}$ on four indicators: best success, final
success, best return, and a composite ranking score used only for within-sweep model
selection (Table~\ref{tab:alpha}). The landscape is \emph{non-monotonic}
(Figure~\ref{fig:alpha}): very small smoothing ($\alpha=0.01$) under-adapts and is
still improving at the horizon, while intermediate values ($\alpha=0.1,0.3$) exhibit
severe late-stage regression despite respectable peaks. Only $\alpha=0.2$ is jointly
best on all four indicators, with a mere 2\,pp peak-to-final gap, so it is fixed for
the canonical QMIX experiments and the exploratory VDN checks. The sharp contrast
between neighbouring values --- $\alpha=0.1$ ends at 42.0\% while $\alpha=0.2$ ends
at 86.0\% --- points to a sensitivity that a single seed cannot characterise, so a
full multi-seed, cross-environment $\alpha$-sweep remains future work; this
diagnostic justifies the fixed value without claiming global optimality.

\begin{table}[!tbp]
  \caption{Diagnostic EMA $\alpha$ ablation on Simple Spread (single seed, 100k
  episodes). Best row in bold.}
  \label{tab:alpha}
  \centering
  \begin{tabular}{@{}lcccc@{}}
    \toprule
    $\alpha$ & Best success rate & Final success rate & Best return & Composite score \\
    \midrule
    0.01 & 76.5\% & 76.0\% & $-30.86$ & 0.671 \\
    0.05 & 82.0\% & 75.0\% & $-32.51$ & 0.480 \\
    0.10 & 63.5\% & 42.0\% & $-32.79$ & 0.150 \\
    \textbf{0.20} & \textbf{88.0\%} & \textbf{86.0\%} & $\mathbf{-29.88}$ & \textbf{0.922} \\
    0.30 & 85.5\% & 57.5\% & $-31.45$ & 0.592 \\
    0.50 & 85.0\% & 85.0\% & $-31.81$ & 0.721 \\
    \bottomrule
  \end{tabular}
\end{table}

\subsection{Overview of the Main Results}\label{sec:overview}

\begin{figure}[!tbp]
  \centering
  \includegraphics[width=0.95\textwidth]{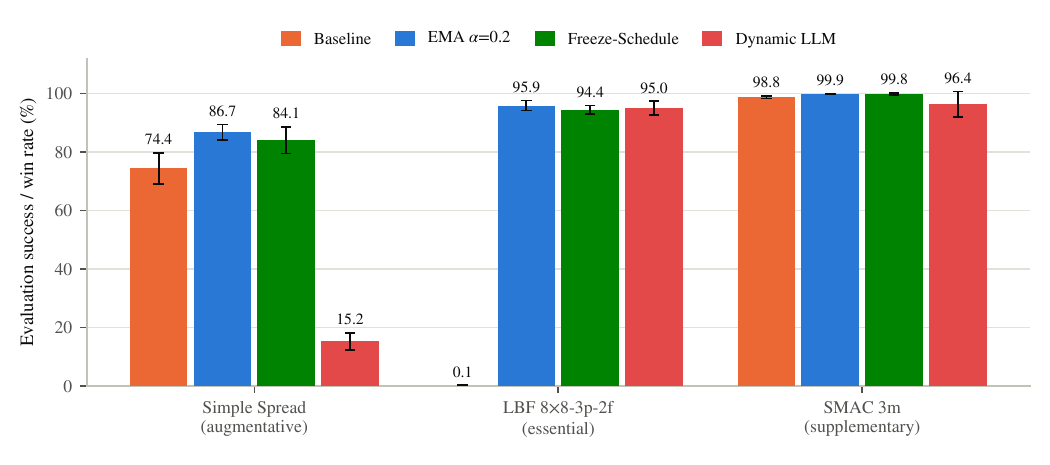}
  \caption{Main results: evaluation success rate (Simple Spread, LBF) and win rate
  (SMAC 3m) for all four conditions, five-seed mean with $\pm1\sigma$ error bars
  (QMIX, $n=5$ seeds). Exact values in Table~\ref{tab:mainresults},
  Appendix~\ref{app:tables}.}
  \label{fig:mainresults}
\end{figure}

Figure~\ref{fig:mainresults} summarises the five-seed comparison across all three
environments and all four conditions; the exact numerical values are collected in
Table~\ref{tab:mainresults} of Appendix~\ref{app:tables}. Three patterns are visible
at a glance and are unpacked in the subsections that follow: on Simple Spread the two
stabilised methods beat the baseline while Dynamic LLM collapses; on LBF every
shaping variant unlocks an otherwise broken task; and on SMAC 3m all methods sit near
the ceiling, with Dynamic LLM alone showing inflated variance.

\subsection{Simple Spread: the Augmentative Regime}\label{sec:spread}

\begin{figure}[!tbp]
  \centering
  \includegraphics[width=0.8\textwidth]{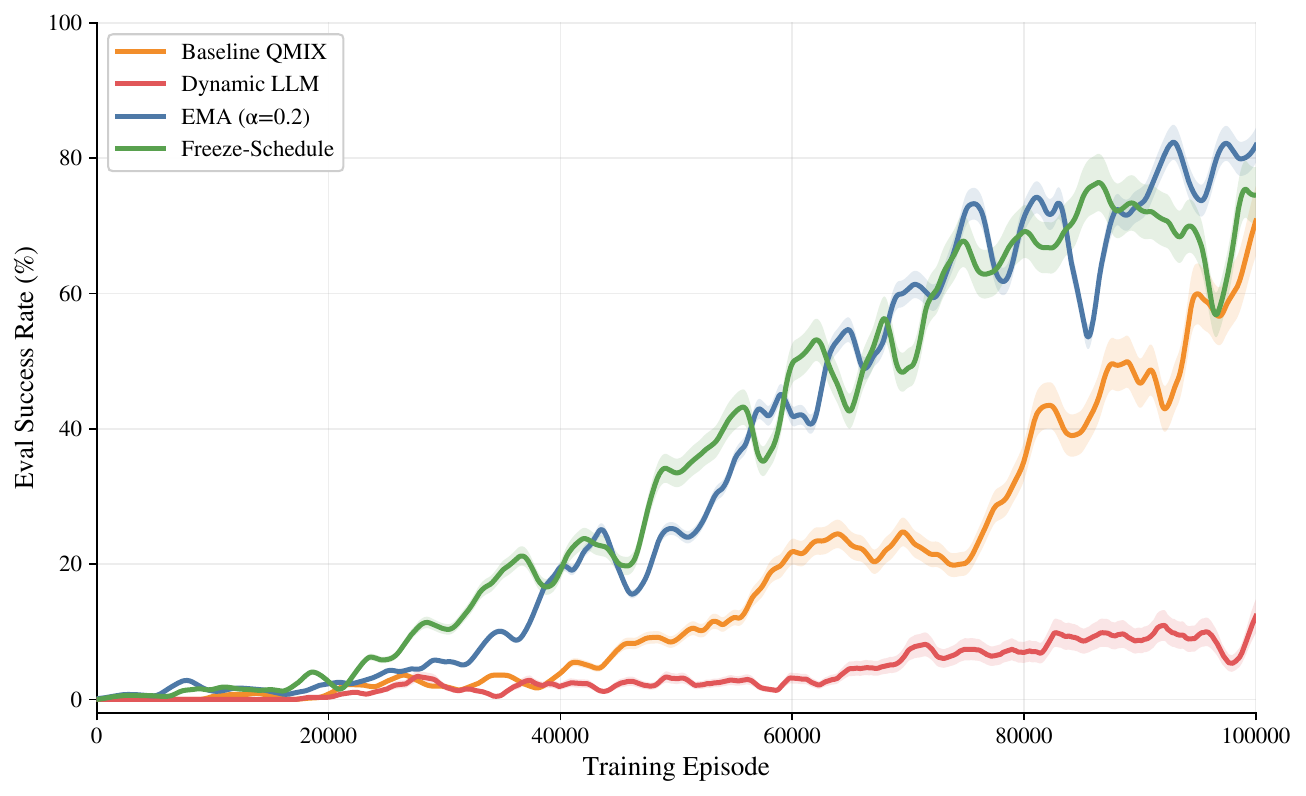}
  \caption{Simple Spread training curves, five seeds. Shaded bands denote $\pm1\sigma$.
  Dynamic LLM (per-episode updates) collapses, whereas EMA and Freeze-Schedule both
  exceed the unshaped baseline.}
  \label{fig:spread}
\end{figure}

Simple Spread requires $N=3$ agents to each cover a distinct landmark while avoiding
collisions. The unshaped baseline is already \emph{functional} (74.4\%), making this
an augmentative regime in which shaping supplements an existing learning signal.
The left group of Figure~\ref{fig:mainresults} reports the five-seed comparison and
Figure~\ref{fig:spread} the training curves.

\begin{figure}[!tbp]
  \centering
  \includegraphics[width=0.85\textwidth]{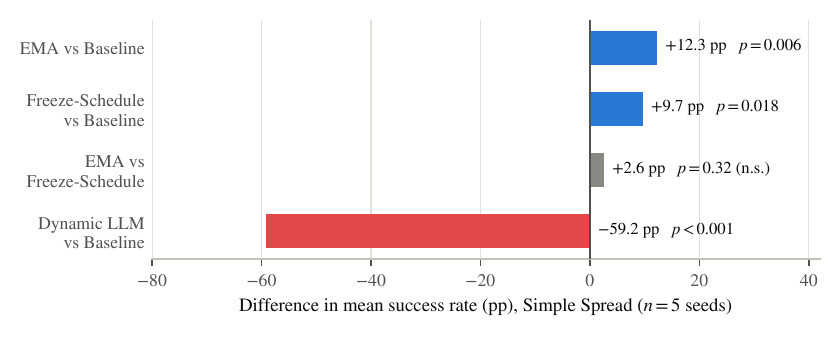}
  \caption{Pairwise Welch $t$-test outcomes on Simple Spread ($n=5$ seeds per
  condition). Bars give the difference in mean success rate; blue marks significant
  gains, red the significant collapse, and grey a non-significant difference. Full
  test statistics in Table~\ref{tab:welch}, Appendix~\ref{app:tables}.}
  \label{fig:welch}
\end{figure}

Both stabilised methods significantly outperform the baseline: EMA reaches
$86.7\pm2.7$ ($+12.3$\,pp, $p<0.01$, Welch $t$-test) and Freeze-Schedule
$84.1\pm4.6$ ($+9.7$\,pp, $p<0.05$). Notably, both exhibit \emph{lower} cross-seed
variance than the baseline itself ($\sigma=2.7$ for EMA, $4.6$ for Freeze, versus
$5.4$ for baseline), indicating that stationarity control also acts as a training
regulariser. In sharp contrast, Dynamic LLM collapses to $15.2\pm3.0$, a fall of
59.2\,pp below baseline ($p<0.001$). This is the defining hazard of the augmentative
regime: because shaping only supplements an already-competent policy, the
non-stationarity noise outweighs the shaping benefit. The pairwise Welch $t$-tests,
visualised in Figure~\ref{fig:welch} (full statistics in Table~\ref{tab:welch},
Appendix~\ref{app:tables}), confirm both stabilised gains are significant while the
two stabilisers are statistically indistinguishable from each other.

\subsection{Level-Based Foraging: the Essential Regime}\label{sec:lbf}

\begin{figure}[!tbp]
  \centering
  \includegraphics[width=0.8\textwidth]{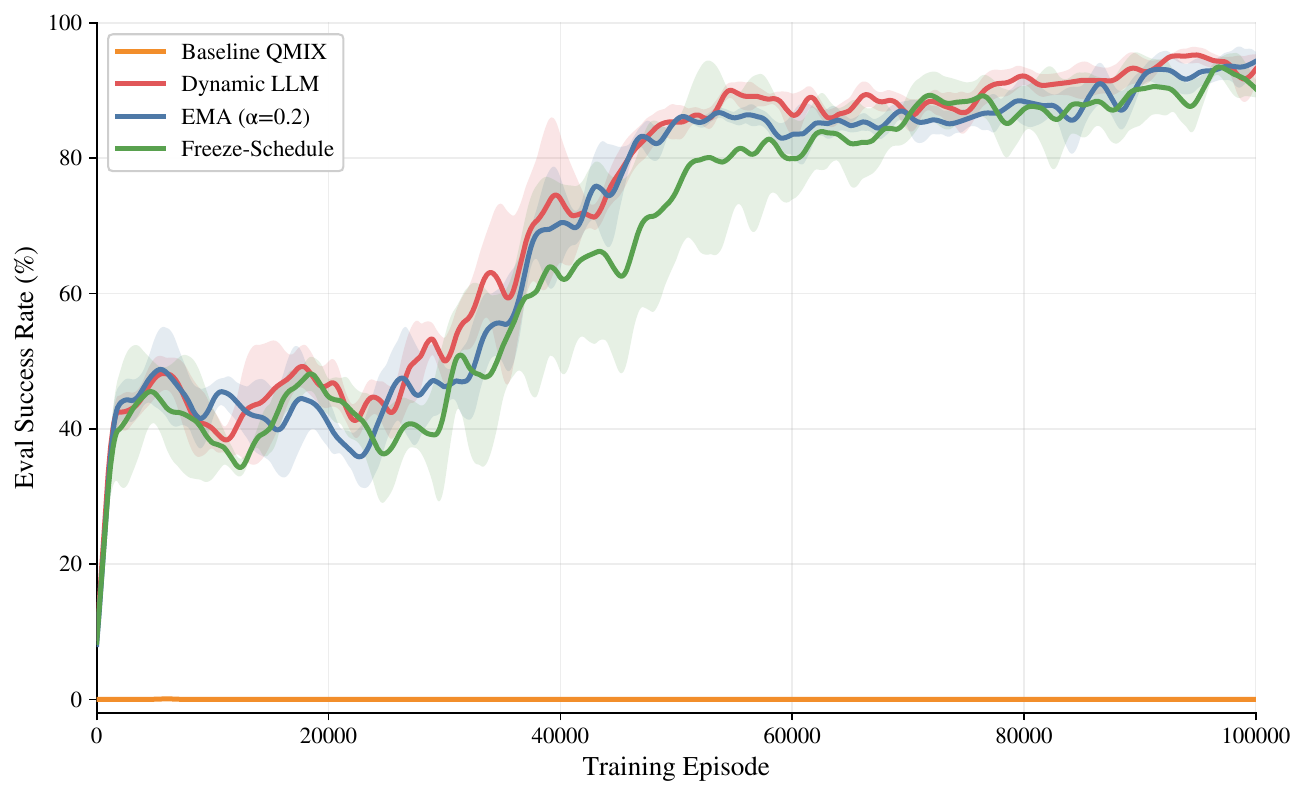}
  \caption{LBF 8$\times$8-3p-2f training curves, five seeds. The unshaped baseline
  never discovers joint-load success; all three shaping variants unlock the task.}
  \label{fig:lbf}
\end{figure}

Level-Based Foraging (8$\times$8-3p-2f) places three agents on a grid where
collecting food requires \emph{simultaneous} loading by enough agents that their
combined levels exceed the food level. The cooperative behaviour is hidden behind a
sparse mechanic that the unshaped policy almost never discovers within the training
budget, making this an \emph{essential} regime in which shaping provides the only
viable learning signal. The centre group of Figure~\ref{fig:mainresults} and
Figure~\ref{fig:lbf} report the results.

The unshaped baseline is operationally broken, succeeding on only $0.1\pm0.2$\% of
episodes: without shaping, the level-sum coordination requirement is intractable
within 100k episodes. All three shaping variants raise success above 94\%, a gain of
$+95.8$\,pp under EMA. The defining feature of the essential regime is that
\emph{even Dynamic LLM works} ($95.0\pm2.4$\%): when the shaping signal is the sole
route to any reward, its benefit dwarfs the non-stationarity noise that proved
catastrophic on Simple Spread. EMA again attains the highest mean and the lowest
cross-seed variance among the shaping methods.

A caveat on attribution is in order. The essential-regime gain demonstrates the
power of PBRS with informative features on a sparse-reward task --- a result in the
spirit of \citep{ng1999policy} --- rather than anything specific to the LLM: a
hand-set static weight vector over the same features $\bm{\phi}(s)$ would plausibly
unlock the task as well. The LLM's contribution here is the natural-language
interface that produces and revises those weights without manual tuning; a no-LLM
static-weight control that isolates this contribution is an experiment we have not
run and flag as future work.

\subsection{SMAC 3m: the Supplementary Regime}\label{sec:smac}

\begin{figure}[!tbp]
  \centering
  \includegraphics[width=0.8\textwidth]{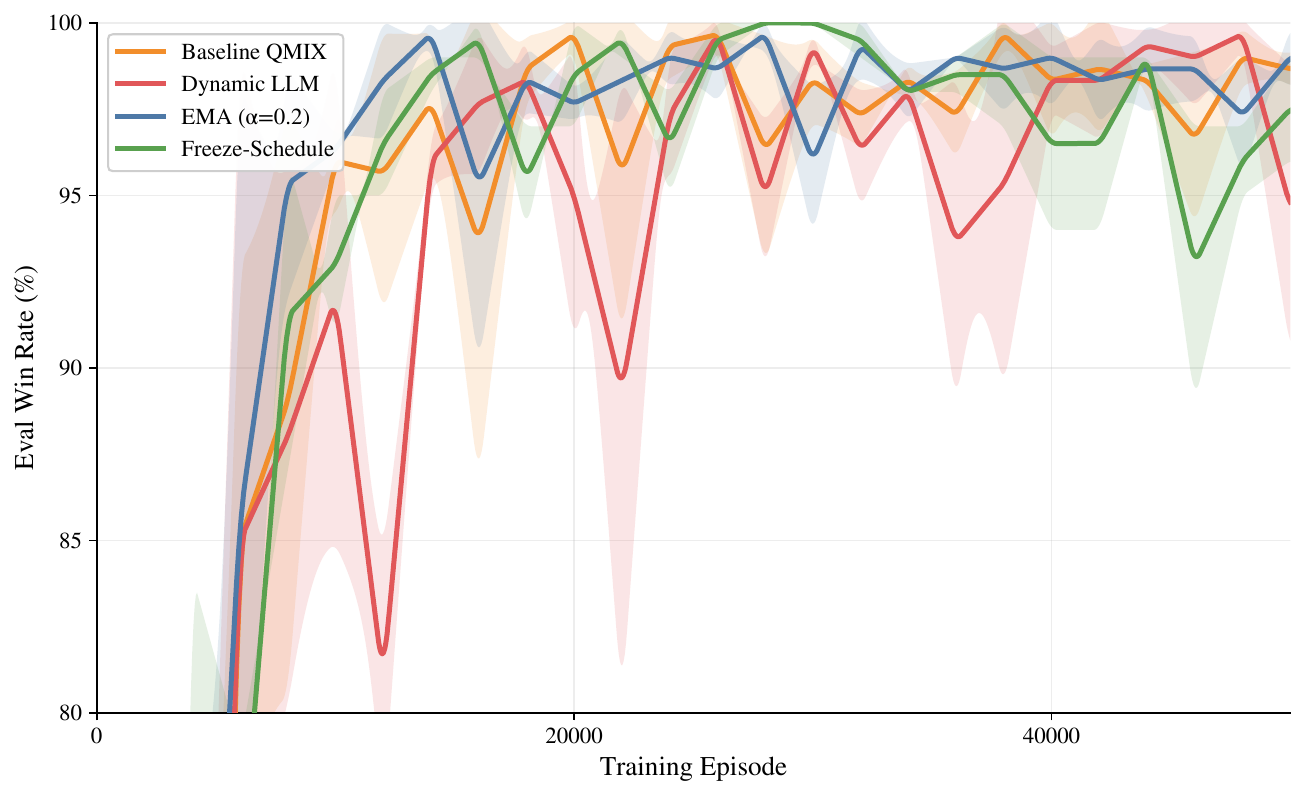}
  \caption{SMAC 3m training curves, shown with a zoomed 80--100\% $y$-axis to expose
  post-convergence volatility; note that the truncated axis visually magnifies
  differences that are small in absolute terms. Win rates are near-ceiling; Dynamic
  LLM shows the largest fluctuations.}
  \label{fig:smac}
\end{figure}

SMAC 3m is a 3-versus-3 StarCraft II micromanagement task in which three allied
Marines must defeat three enemy Marines through focused fire and positioning. We use
the SMAClite backend, a lightweight reimplementation of SMAC; absolute win rates are
therefore not directly comparable to published results on the original benchmark,
and comparisons should once more be made within the environment. Baseline QMIX is
already near-saturated, so shaping cannot
meaningfully raise the mean; the regime is \emph{supplementary}, and the relevant
test is whether stabilisation \emph{avoids harming} an already-strong policy.
The right group of Figure~\ref{fig:mainresults} and Figure~\ref{fig:smac} report
the results.

With a baseline of $98.8\pm0.5$\%, the EMA and Freeze gains ($+1.1$\,pp and
$+1.0$\,pp) are not statistically significant --- there is simply no headroom. The
informative signal is in the tail: Dynamic LLM drops to $96.4\pm4.3$\%, adding
substantial variance without any mean gain, while EMA and Freeze hold at 99.9\% and
99.8\% with negligible spread. In the supplementary regime, stabilised shaping passes
the do-no-harm test that unstabilised dynamic updates fail.

\subsection{Per-Seed Breakdown}\label{sec:perseed}

\begin{figure}[!tbp]
  \centering
  \includegraphics[width=0.95\textwidth]{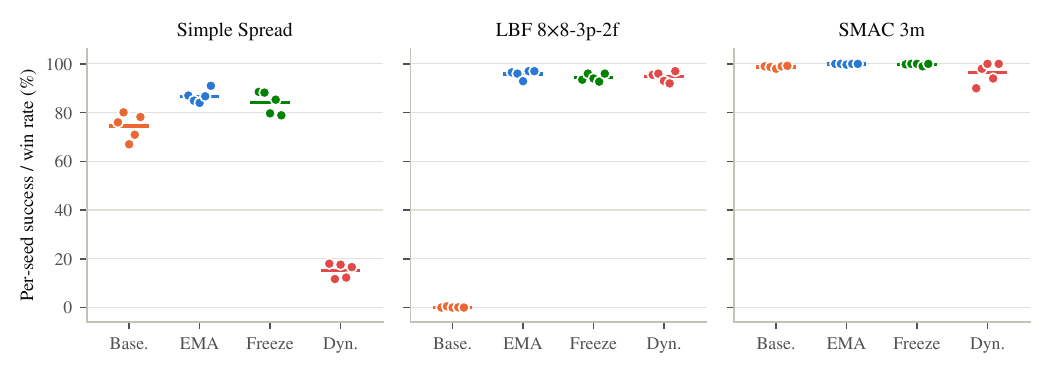}
  \caption{Per-seed evaluation outcomes for every environment and method (seeds
  42/101/123/456/999); each dot is one seed and the horizontal bar the five-seed
  mean. Numerical values in Table~\ref{tab:perseed}, Appendix~\ref{app:tables}.}
  \label{fig:perseed}
\end{figure}

Figure~\ref{fig:perseed} plots the individual seed outcomes underlying the aggregate
means above (numerical values in Table~\ref{tab:perseed}, Appendix~\ref{app:tables}).
The Dynamic LLM collapse on Simple Spread is consistent across all five seeds (range
11.7--18.0\%), ruling out a seed-specific artefact, and the LBF baseline rounds to
zero on every seed.

\subsection{Exploratory VDN Results}\label{sec:vdn}

A natural concern is that the regime taxonomy might be an artefact of QMIX's
monotonic mixing network rather than a property of the environments themselves. To
probe this, we ran an exploratory extension in which the mixer is replaced by VDN's
simpler additive decomposition \citep{sunehag2018vdn}, while the backbone, training
budget, shaping pipeline, and evaluation protocol all remain unchanged.

On two of the three environments, VDN reproduces the taxonomy. On Simple Spread, the
ordering of conditions matches the QMIX study: EMA achieves the best mean
($81.6\pm4.3$\%), slightly above the unshaped baseline ($80.0\pm10.9$\%), while
Dynamic LLM is both the weakest and the most variable condition ($75.2\pm10.8$\%).
The collapse is milder than under QMIX, but the augmentative signature ---
stabilised shaping helps, unstabilised shaping hurts --- is preserved. On SMAC 3m,
VDN reproduces the supplementary ceiling: the baseline, EMA, and every completed
Dynamic LLM seed reach a 100\% win rate, leaving no headroom for shaping to add or
remove.

LBF is the exception: the shaped VDN runs did not complete reliably. The baseline
reached $49.6\pm25.3$\%, only a single EMA seed produced a usable result (90.0\%),
and every Dynamic LLM seed failed. With so few completed runs, we draw no conclusion
about the essential regime under VDN.

In summary, the VDN extension provides preliminary supporting evidence for the
taxonomy on Simple Spread and SMAC 3m, but it is not an independent confirmation;
every regime-level claim in this paper is anchored in the five-seed QMIX study.

\subsection{The Regime Taxonomy}\label{sec:taxonomy}

\begin{figure}[!tbp]
  \centering
  \includegraphics[width=0.95\textwidth]{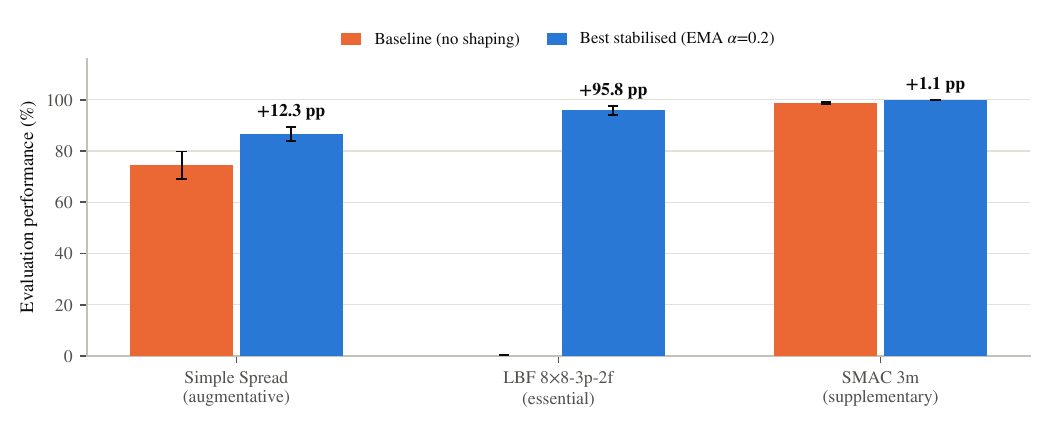}
  \caption{Cross-environment comparison underlying the regime taxonomy (QMIX, $n=5$
  seeds). Bars show mean $\pm1\sigma$; the annotated deltas give the best stabilised
  gain over the unshaped baseline in each regime.}
  \label{fig:crossenv}
\end{figure}

Synthesising the three environments yields a taxonomy organised by how competent the
unshaped baseline already is (Figure~\ref{fig:crossenv}; the placements are tabulated
in Table~\ref{tab:taxonomy}, Appendix~\ref{app:tables}).

The three regimes are defined by baseline competence and predict the value --- and
danger --- of dynamic LLM shaping. In the \emph{essential} regime (LBF), the baseline
is broken and any shaping unlocks the task, so even non-stationary updates help. In
the \emph{augmentative} regime (Simple Spread), the baseline is competent and shaping
only supplements it; controlled shaping adds significant gains, but unstabilised
dynamic updates collapse the policy. In the \emph{supplementary} regime (SMAC 3m),
the baseline is near-saturated and shaping is largely redundant; stabilised shaping
is harmless, while unstabilised shaping injects variance. Regime placement, then, is a
practical predictor of whether to apply dynamic LLM shaping and how aggressively to
stabilise it.

An important qualification bounds this claim. With a single exemplar per regime,
baseline competence is necessarily confounded with every other property that
separates the three environments --- reward density, horizon, observation
dimensionality, and cooperation mechanic (Table~\ref{tab:envs}). The taxonomy should
therefore be read as an organising hypothesis that is consistent with, and motivated
by, the three case studies, not as an established predictive law. Populating each
regime with additional environments, or sweeping a single environment across regimes
(for instance, varying LBF's grid size and food levels moves its baseline from
broken to functional), is the direct test we identify for future work.

\section{Interpretability and Human-in-the-Loop Evaluation}\label{sec:interp}

Because the framework is meant to keep a human in the loop, reward control is paired
with an interpretability subsystem: an \emph{EpisodeCollector} records actions,
Q-values, and task events; a \emph{MomentDetector} flags decision-critical moments
via per-environment Q-margin thresholds; and a \emph{BehaviorExplainer} renders these
into natural-language audit narratives. Applied to the best checkpoint of each
environment over 100 evaluation episodes, the same three-component pipeline produces
coherent narratives across all three regimes (Table~\ref{tab:explain}). The mean
Q-margin grows by an order of magnitude across environments
($0.027\to0.052\to0.156$), motivating per-environment threshold calibration; every
episode received an LLM-generated narrative without fallback, at $\approx 1.91$\,GB
GPU memory and $\approx 5$ minutes per 100 episodes. Pairing each narrative with the
structured \texttt{KeyMoment} ground truth makes the explanations auditable --- a
transparency prerequisite for deployment. The evaluations in this section re-load
saved checkpoints and run independent 100--200-episode evaluations, so absolute rates
differ by a few points from the best-checkpoint figures of Section~\ref{sec:experiments}.

\begin{table}[!tbp]
  \caption{Cross-environment explainability profile ($n=100$ episodes per environment).}
  \label{tab:explain}
  \centering
  \begin{tabular}{@{}llll@{}}
    \toprule
    Property & Simple Spread & LBF 8$\times$8-3p-2f & SMAC 3m \\
    \midrule
    Checkpoint            & EMA, seed 42 & EMA, seed 42 & EMA, seed 456 \\
    Avg. episode length   & 25.0 / 25 & 48.7 / 50 & 19.0 \\
    Avg. key moments / ep. & 8.66 & 5.28 & 12.00 \\
    Q-margin threshold $\tau_Q$ & 0.30 & 0.50 & 1.00 \\
    Mean Q-margin         & 0.027 & 0.052 & 0.156 \\
    Top moment type       & coverage & near\_food & focus\_fire \\
    LLM narrative coverage & 100\% & 100\% & 100\% \\
    \bottomrule
  \end{tabular}
\end{table}

\subsection{Frozen-Policy Instruction Evaluation}\label{sec:frozen}

On the frozen EMA seed-42 Spread checkpoint, ten operator instructions spanning
speed, safety, and coordination intent were translated to weight vectors and each
evaluated over 100 episodes. Success and collision rates were \emph{identical} across
all ten (SR $=0.85$, CR $=0.22$): with a frozen Q-network and an observation that
excludes the instruction, action selection $\arg\max_a Q(\tau,u)$ cannot change. The
shaped returns, however, differ per instruction (range $-31.20$ to $-30.43$ over an
identical base return $-31.74$), confirming that the LLM produces semantically
distinct $\bm{w}$. We report this honestly as instruction-aware \emph{shaped-reward}
evaluation, not online behavioural adaptation.

\subsection{Post-Instruction Fine-Tuning}\label{sec:posthoc}

To test whether \emph{unfreezing} the policy yields behavioural specialisation, we
locked the weights to three contrasting instructions and continued training for 5k
episodes, with learning rate $\eta=10^{-4}$ and exploration rate $\varepsilon$
annealed from 0.10 to 0.01. All three collapsed
(Table~\ref{tab:finetune}). This honest negative result reinforces the central
message: late-stage reward re-specification destabilises an off-policy learner
exactly as mid-training non-stationarity does. It argues for evaluation-time
interpretability and human override --- rather than weight retraining --- as the safe
mode of human-in-the-loop control.

\begin{table}[!tbp]
  \caption{Post-instruction fine-tuning (5k episodes per instruction, frozen EMA s42
  checkpoint). $\bm{w}=(w_\ell,w_g,w_c)$.}
  \label{tab:finetune}
  \centering
  \begin{tabular}{@{}lcccc@{}}
    \toprule
    Instruction & $\bm{w}$ & Pre SR & Post SR & $\Delta$SR \\
    \midrule
    Rush to landmark & $(5,0,1)$ & 80.0\% & 0.0\%  & $-80.0$ \\
    Avoid collisions & $(0,2,5)$ & 80.0\% & 0.5\%  & $-79.5$ \\
    Team coordination & $(1,5,2)$ & 80.0\% & 0.0\% & $-80.0$ \\
    \bottomrule
  \end{tabular}
\end{table}

\section{Discussion}\label{sec:discussion}

\paragraph{A stability--quality paradox.} The LLM-reward literature focuses on the
\emph{semantic quality} of generated rewards: Eureka searches evolutionary
generations of reward programs for better-performing candidates, and Text2Reward
refines reward code from human feedback, implicitly treating each revision as an
improvement to be applied as soon as it is available. Our results show that in
off-policy MARL the \emph{stability} of the reward signal matters as much as its
quality: even a well-designed reward causes catastrophic collapse if it is updated
faster than the agent can adapt (Simple Spread, $74.4\%\to15.2\%$). The paradox is
that a higher-quality revision, applied at the wrong cadence, is strictly worse
than no revision at all. Reward-signal stationarity is therefore a first-class
design constraint, not an implementation detail: any system that couples an LLM
reward designer to an off-policy learner should specify an update budget --- how
often weights may change, and through what smoothing --- with the same care it
devotes to prompt design.

\paragraph{Stationarity--adaptability trade-off.} The two stabilisers occupy
different points on a continuum. The Freeze-Schedule enforces strict stationarity
within each phase --- exact PBRS invariance, at the price of discontinuities at
phase boundaries --- while EMA admits a small, bounded drift at every episode,
spreading each weight revision over roughly $1/\alpha$ episodes. That EMA
(quasi-stationarity) slightly outperforms the Freeze-Schedule (strict stationarity)
on Simple Spread suggests that moderate, controlled drift can act as an implicit
curriculum rather than pure noise: the potential is always moving gently towards
the operator's intent, and the learner tracks it. The advantage is small and
environment-dependent, but it indicates that the design goal is bounded drift, not
its complete elimination --- and that the smoothing factor $\alpha$ is the natural
knob with which to position a system on this continuum.

\paragraph{Lightweight LLMs as reward architects.} Constraining the LLM to emit a
structured JSON weight vector within a fixed PBRS potential reduces the task to
translation, so a 3B model suffices at $<200$\,ms per query with negligible compute.
This is a deliberate trade against expressiveness: code-generating approaches such
as Eureka and Text2Reward let the LLM write arbitrary reward programs, which is
strictly more powerful but harder to audit and can introduce unbounded or
non-potential terms that void the PBRS guarantee by construction. The weight-vector
interface keeps every degree of freedom semantically meaningful and bounded ---
each $w_j$ answers ``how much does the operator currently care about feature
$\phi_j$?'' --- which the interpretability subsystem of Section~\ref{sec:interp}
exploits directly. Practically, the small footprint ($\approx1.91$\,GB) means the
reward architect runs on the same consumer GPU as training, with no external API
dependency, which matters for on-premise and latency-sensitive deployments.

\paragraph{Future directions.} The present study anchors the regime taxonomy in a
single LLM, three environments, and small teams, which opens several natural
extensions. First, sharpening the taxonomy's evidential basis: populating each
regime with at least two environments, or sweeping a single environment across
regimes (varying LBF's grid size and food levels moves its baseline from broken to
functional at modest cost), would disentangle baseline competence from the other
properties on which the three environments differ. Second, converting the
mechanistic account of Section~\ref{sec:nonstat} from asserted to demonstrated: the
account predicts that intermediate update frequencies without smoothing (e.g., every
1k--5k episodes) interpolate between the Dynamic and Freeze-Schedule endpoints, and
that collapse severity scales with replay-buffer size --- both are directly testable
causal predictions. Third, isolating the LLM's contribution with a static hand-set
weight baseline over the same features, and testing robustness to the choice of
reward architect beyond Qwen 2.5 3B. Fourth, establishing algorithm-agnosticism by
completing the VDN sweeps on LBF and extending to policy-gradient methods such as
MAPPO and richer decompositions such as QPLEX, together with a full multi-seed,
cross-environment sweep of the smoothing factor $\alpha$
(Section~\ref{sec:alpha}). Finally, the observation that bounded weight drift can
act as an implicit curriculum invites deliberate curriculum design over the shaping
weights, paired with real-time human-in-the-loop evaluation built on the
interpretability stack of Section~\ref{sec:interp}.

\section{Conclusion}\label{sec:conclusion}

Per-episode LLM weight updates break the stationarity assumption of potential-based
reward shaping and collapse task performance in the augmentative regime
($74.4\%\to15.2\%$ on Simple Spread). Two stabilisation strategies restore it: a
Phase-Based Freeze Schedule and EMA smoothing ($\alpha=0.2$), the latter reaching
86.7\% on Simple Spread, 95.9\% on Level-Based Foraging, and 99.9\% on SMAC 3m.
Across three environments and five seeds with QMIX --- with exploratory VDN support
--- the results motivate a three-regime taxonomy (essential, augmentative,
supplementary) that predicts when dynamic LLM shaping helps, when it is dangerous,
and when it is merely redundant; with one environment per regime, we advance it as
an organising hypothesis whose predictive scope remains to be established. The
broader lesson is that reward-signal stationarity is a necessary design constraint
for any system that combines LLM-guided reward shaping with off-policy experience
replay. Future work will test the mechanism's causal predictions (intermediate
update frequencies and buffer-size ablations), add static-weight and multi-LLM
controls, populate each regime with further environments, complete the VDN sweeps,
extend to MAPPO and QPLEX, and develop real-time human-in-the-loop evaluation.

\vspace{5mm}
\noindent
\textbf{Acknowledgements}\\
This work was carried out partially using the AI Datacenter of the National School of
Artificial Intelligence, funded under grant number E049 24 0117 by the Algerian
Ministry of Higher Education and Scientific Research.

\vspace{5mm}
\noindent
\textbf{Declarations}
\begin{itemize}
  \item \textbf{Funding.} This work was supported by the Algerian Ministry of Higher
  Education and Scientific Research under grant number E049 24 0117 (AI Datacenter,
  National School of Artificial Intelligence).
  \item \textbf{Competing interests.} The authors declare that they have no competing
  interests.
  \item \textbf{Ethics approval and consent to participate.} Not applicable.
  \item \textbf{Consent for publication.} Not applicable.
  \item \textbf{Data availability.} The training logs and experimental
  configurations supporting the findings of this paper will be released in a public
  repository upon acceptance; until then they are available from the authors upon
  reasonable request.
  \item \textbf{Materials availability.} Not applicable.
  \item \textbf{Code availability.} The full codebase will be released in a public
  repository upon acceptance; until then it is available from the authors upon
  reasonable request.
  \item \textbf{Author contribution.} F.~Keddouri and S.~Houhou designed and
  implemented the framework, ran the experiments, and drafted the manuscript.
  A.~Boulmerka and N.~Farhi supervised the work, advised on methodology and
  evaluation, and revised the manuscript. All authors read and approved the final
  manuscript.
\end{itemize}

\bibliographystyle{plainnat} 
\bibliography{references}

\clearpage

\appendix

\section{Numerical Result Tables}\label{app:tables}

This appendix collects the numerical values underlying the result figures of
Section~\ref{sec:experiments}. Table~\ref{tab:mainresults} merges the per-environment
five-seed comparisons visualised in Figure~\ref{fig:mainresults};
Table~\ref{tab:welch} gives the full Welch $t$-test statistics behind
Figure~\ref{fig:welch}; Table~\ref{tab:perseed} lists the per-seed outcomes plotted
in Figure~\ref{fig:perseed}; and Table~\ref{tab:taxonomy} tabulates the regime
placements of Figure~\ref{fig:crossenv}.

\begin{table}[htbp]
  \caption{Five-seed comparison across all three environments (QMIX, $n=5$, seeds
  42/101/123/456/999; mean$\pm$std). Eval success rate (\%) for Simple Spread and
  LBF, win rate (\%) for SMAC 3m. Best method per environment in bold.}
  \label{tab:mainresults}
  \centering
  \begin{tabular}{@{}lccc@{}}
    \toprule
    & Simple Spread & LBF 8$\times$8-3p-2f & SMAC 3m \\
    Method & SR (\%) & SR (\%) & WR (\%) \\
    \midrule
    Baseline QMIX (no shaping) & $74.4\pm5.4$ & $0.1\pm0.2$ & $98.8\pm0.5$ \\
    EMA $\alpha=0.2$           & $\mathbf{86.7\pm2.7}$ & $\mathbf{95.9\pm1.7}$ & $\mathbf{99.9\pm0.1}$ \\
    Freeze-Schedule            & $84.1\pm4.6$ & $94.4\pm1.5$ & $99.8\pm0.4$ \\
    Dynamic LLM (per-episode)  & $15.2\pm3.0$ & $95.0\pm2.4$ & $96.4\pm4.3$ \\
    \bottomrule
  \end{tabular}
\end{table}

\begin{table}[htbp]
  \caption{Pairwise Welch $t$-tests on Simple Spread ($n=5$ seeds per condition).
  Here $t$ is the Welch test statistic, df the Welch--Satterthwaite degrees of
  freedom, $p$ the two-sided $p$-value, and $\Delta$ the difference in mean success
  rate in percentage points (pp). The Welch test is used because the collapsed
  Dynamic LLM variance is incommensurable with the stabilised methods'.}
  \label{tab:welch}
  \centering
  \begin{tabular}{@{}lccccl@{}}
    \toprule
    Comparison & $t$ & df & $p$ & $\Delta$ (pp) & Verdict \\
    \midrule
    EMA vs Baseline           & 4.56  & 5.9 & 0.006 & $+12.3$ & Significant ($p<0.01$) \\
    Freeze-Schedule vs Baseline & 3.06 & 7.8 & 0.018 & $+9.7$  & Significant ($p<0.05$) \\
    EMA vs Freeze-Schedule    & 1.09  & 6.5 & 0.32  & $+2.6$  & Not significant \\
    Dynamic LLM vs Baseline   & $-21.4$ & 6.3 & $<0.001$ & $-59.2$ & Highly significant (collapse) \\
    \bottomrule
  \end{tabular}
\end{table}

\begin{table}[htbp]
  \caption{Per-seed results (\%) for all environments and methods (seeds
  42/101/123/456/999). SR for Spread/LBF, WR for SMAC.}
  \label{tab:perseed}
  \centering
  \begin{tabular}{@{}llcccccc@{}}
    \toprule
    Env & Method & s42 & s101 & s123 & s456 & s999 & Mean$\pm$std \\
    \midrule
    \multirow{4}{*}{Spread}
      & Baseline       & 76.0 & 80.1 & 67.0 & 70.9 & 78.2 & $74.4\pm5.4$ \\
      & EMA $\alpha=0.2$ & 87.0 & 84.9 & 84.0 & 86.7 & 91.0 & $\mathbf{86.7\pm2.7}$ \\
      & Freeze         & 88.5 & 88.2 & 79.7 & 85.3 & 78.9 & $84.1\pm4.6$ \\
      & Dynamic LLM    & 18.0 & 11.7 & 17.6 & 12.3 & 16.6 & $15.2\pm3.0$ \\
    \midrule
    \multirow{4}{*}{LBF}
      & Baseline       & 0.0 & 0.5 & 0.0 & 0.1 & 0.0 & $0.1\pm0.2$ \\
      & EMA $\alpha=0.2$ & 96.5 & 96.0 & 92.9 & 97.0 & 97.0 & $\mathbf{95.9\pm1.7}$ \\
      & Freeze         & 93.5 & 96.0 & 94.0 & 92.7 & 96.0 & $94.4\pm1.5$ \\
      & Dynamic LLM    & 95.5 & 96.0 & 93.0 & 92.0 & 97.0 & $95.0\pm2.4$ \\
    \midrule
    \multirow{4}{*}{SMAC}
      & Baseline       & 99.0 & 98.7 & 98.0 & 99.0 & 99.2 & $98.8\pm0.5$ \\
      & EMA $\alpha=0.2$ & 100.0 & 100.0 & 99.7 & 100.0 & 100.0 & $\mathbf{99.9\pm0.1}$ \\
      & Freeze         & 99.8 & 100.0 & 100.0 & 99.0 & 100.0 & $99.8\pm0.4$ \\
      & Dynamic LLM    & 90.0 & 98.0 & 100.0 & 94.0 & 100.0 & $96.4\pm4.3$ \\
    \bottomrule
  \end{tabular}
\end{table}

\begin{table}[htbp]
  \caption{Environment placement under the three-regime taxonomy (QMIX, $n=5$ seeds).
  ``Best'' is the strongest stabilised method (EMA in all three).}
  \label{tab:taxonomy}
  \centering
  \begin{tabular}{@{}llcc@{}}
    \toprule
    Environment & Regime & Baseline & Best \\
    \midrule
    LBF           & Essential      & 0.1\%  & 95.9\% \\
    Simple Spread & Augmentative   & 74.4\% & 86.7\% \\
    SMAC 3m       & Supplementary  & 98.8\% & 99.9\% \\
    \bottomrule
  \end{tabular}
\end{table}

\end{document}